\documentclass{article}

\usepackage[preprint]{neurips_2025}

\usepackage[utf8]{inputenc} 
\usepackage[T1]{fontenc}    
\usepackage{hyperref}       
\usepackage{url}            
\usepackage{booktabs}       
\usepackage{amsfonts}       
\usepackage{nicefrac}       
\usepackage{microtype}      
\usepackage{xcolor}         

\usepackage{booktabs, threeparttable}
\usepackage{amsfonts}       
\usepackage{nicefrac}       
\usepackage{microtype}      
\usepackage[table,xcdraw, dvipsnames]{xcolor}
\usepackage{graphicx} 
\usepackage{wrapfig}
\usepackage{subcaption}
\usepackage{amsmath}
\usepackage{svg}
\usepackage{diagbox} 

\definecolor{myOrange}{HTML}{EDBC84}

\title{The Quest for Universal Master Key Filters in DS-CNNs}

\author{%
  Zahra Babaiee\thanks{Equal contribution.} \\
  Technische Universit{\"a}t Wien\\
  Vienna, Austria \\
  \texttt{zahra.babaiee@tuwien.ac.at} \\
  \And
  Peyman M. Kiassari\thanks{Equal contribution.} \\
  Technische Universit{\"a}t Wien\\
  Vienna, Austria \\
  \texttt{peyman.kiasari@tuwien.ac.at} \\
  \And
  Daniela Rus \\
  Massachusetts Institute of Technology\\
  Cambridge, MA \\
  \texttt{rus@mit.edu} \\
  \And
  Radu Grosu \\
  Technische Universit{\"a}t Wien\\
  Vienna, Austria \\
  \texttt{radu.grosu@tuwien.ac.at} \\
}

\begin{document}

\maketitle

\begin{abstract}
A recent study has proposed the ``Master Key Filters Hypothesis'' for convolutional neural network filters. This paper extends this hypothesis by radically constraining its scope to a single set of just 8 universal filters that depthwise separable convolutional networks inherently converge to. While conventional DS-CNNs employ thousands of distinct trained filters, our analysis reveals these filters are predominantly linear shifts (ax+b) of our discovered universal set. Through systematic unsupervised search, we extracted these fundamental patterns across different architectures and datasets. Remarkably, networks initialized with these 8 unique frozen filters achieve over 80\% ImageNet accuracy, and even outperform models with thousands of trainable parameters when applied to smaller datasets. The identified master key filters closely match Difference of Gaussians (DoGs), Gaussians, and their derivatives, structures that are not only fundamental to classical image processing but also strikingly similar to receptive fields in mammalian visual systems. Our findings provide compelling evidence that depthwise convolutional layers naturally gravitate toward this fundamental set of spatial operators regardless of task or architecture. This work offers new insights for understanding generalization and transfer learning through the universal language of these master key filters.
\end{abstract}

\section{Introduction}

Convolutional Neural Networks (CNNs) have significantly advanced computer vision through their hierarchical representations using trainable filters. As architectures evolved toward greater performance, models such as VGG~\cite{DBLP:journals/corr/SimonyanZ14a}, ResNet~\cite{he2016residual}, and DenseNet~\cite{DenseNet} incorporated thousands of filters across their layers. This trend continued with the development of Depthwise Separable Convolutional Neural Networks (DS-CNNs)~\cite{mobilenets, mobilenetv3}, which separate spatial and channel-wise computations for improved efficiency. Contemporary architectures like the ConvNeXt family~\cite{convnext, ConvNeXtV2} utilize DS-CNNs with up to 50,000 trainable spatial filters.

Recent research identified a notable pattern in trained depthwise convolutional kernels across various DS-CNN architectures~\cite{babaiee2023unveiling}. Through analysis of trained filters using unsupervised clustering, they demonstrated that these patterns converge into distinct clusters resembling Difference of Gaussian (DoG) functions and their derivatives. Their study classified over 95\% and 90\% of filters from ConvNextV2 and ConvNeXt models, respectively, into Gaussian-related clusters, indicating consistent patterns in filter learning.

Subsequently, another work proposed the ``Master Key Filters Hypothesis,''~\cite{babaiee2024master} proposing that there exist master key filter sets that are general for visual data, and that the depthwise filters in DS-CNNs tend to converge to these master key filters, regardless of the specific dataset, task, or architecture. This hypothesis challenges the conventional understanding that convolutional filters become increasingly specialized in deeper layers and suggests instead that a set of fundamental filters may underlie the performance of these networks.
\begin{figure*}[t]
  \centering
  \includegraphics[width=0.9\linewidth]{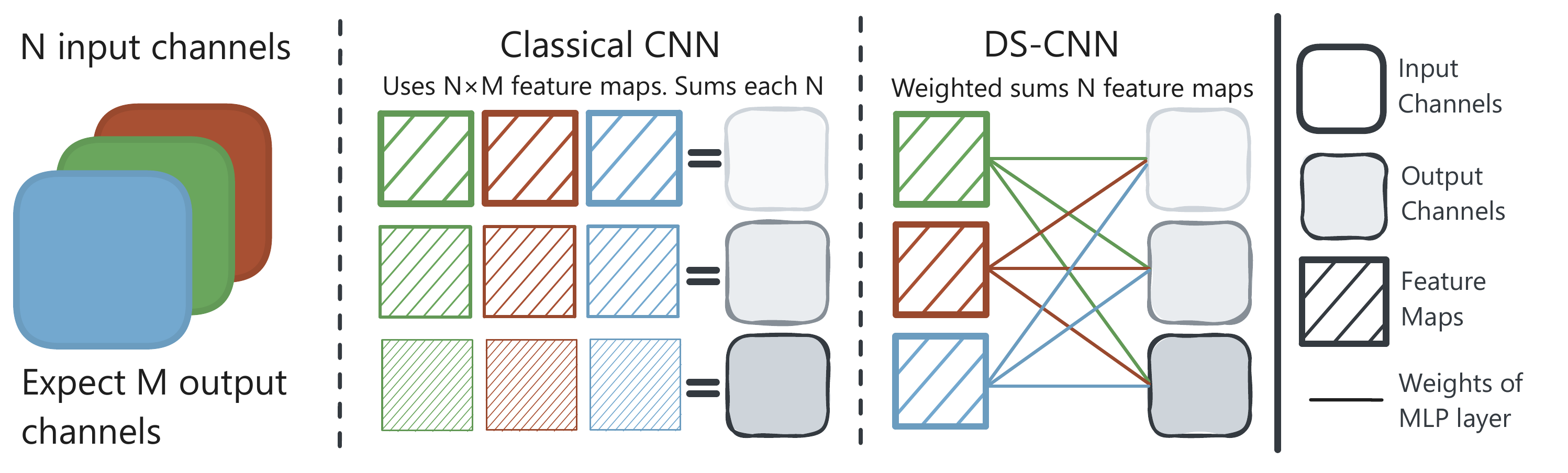}
   \caption{\textbf{Comparison of Classical CNN and DS-CNN architectures:} Left: Input with N channels. Center: In Classical CNNs, each output channel is produced by convolving a unique filter with each input channel, followed by summing the resulting feature maps. This results in N×M distinct filters and corresponding feature maps. Right: DS-CNN uses only N filters (one per input channel) to create N feature maps, then applies an N×M MLP layer to linearly combine these feature maps into M output channels. DS-CNNs represent a parameter-efficient \emph{subset} of classical CNNs, reducing the number of required convolutional operations.}
   \label{fig:dscnn}
\end{figure*}
In this paper, we extend the "Master Key Filters Hypothesis" by radically constraining its scope through identification of a minimal fundamental master key filter set. While the original hypothesis posited the existence of general-purpose filter sets for visual data—potentially comprising numerous filters across multiple sets—our systematic unsupervised analysis across architectures and datasets reveals a remarkably compact representation. We demonstrate that DS-CNNs predominantly converge toward a basis of just 8 distinct filters, where a substantial proportion of learned filters approximate linear shifts of these fundamental kernels. Notably, networks restricted to this compact basis maintain performance integrity, suggesting these filters capture essential visual processing primitives rather than task-specific features. This finding significantly refines the original hypothesis by establishing both the cardinality and specific form of a universal filter basis for visual computing.

Our work also substantially refines the observations made in ~\cite{babaiee2023unveiling}, which identified Gaussian-related patterns in DS-CNN filters without constraining their potential variability. While that study demonstrated the prevalence of Guassian-like filters, it allowed for an effectively infinite continuum of these structures with arbitrary standard deviations and noise characteristics. In contrast, we're narrowing it down as linear shift of a mere 8 fundamental filters. This significantly narrows the theoretical space.

These identified filters correspond to mathematical forms matching Difference of Gaussians (DoGs), Gaussians, and their derivatives, which are established components in scale-space theory~\cite{lindeberg2013scale} and share structural similarities with receptive fields observed in mammalian visual systems~\cite{young2001gaussian, YoungDoG}. Networks initialized with these 8 filters achieve over 80\% ImageNet accuracy and demonstrate superior performance compared to models with thousands of trainable parameters when applied to smaller datasets.

Our findings provide empirical support for the ``Master Key Filters Hypothesis'' and suggest potential applications in efficient network design and transfer learning, while contributing to the understanding of generalizable patterns in visual processing systems.

\section{Related Work}

\textbf{Depthwise Convolutional Filters.} Depthwise Convolutions (DCs) have revolutionized the design of Convolutional Neural Networks (CNNs) by using only one feature map per input channel, leading to the development of lightweight and performant architectures such as MobileNet~\cite{mobilenets}, EfficientNet~\cite{efficientnet}, and ConvNeXt~\cite{convnext}. As illustrated in Figure~\ref{fig:dscnn}, classical CNNs utilize $c_{in} \times c_{out}$ completely separate filters, creating independent feature maps for each input-output channel combination. Each output channel is computed as: 
$$Y_i = \sum_{j=1}^{c_{in}} K_{i,j} \ast X_j, \text{for } (i, j) \in [c_{out}] \times [c_{in}]$$
where $X \in \mathbb{R}^{H\times W\times c_{in}}$ is the input tensor with $c_{in}$ channels, $Y_i$ is the $i$-th output channel, $K_{i,j}$ is the convolutional kernel for the $i$-th output channel and $j$-th input channel, and $[n]$ denotes the set $\{1,2,...,n\}$.

In contrast, DS-CNNs force the model to use only $c_{in}$ feature maps (one per input channel) followed by linearly combining these feature maps using pointwise convolutions or MLP layers. It is important to emphasize that DS-CNNs combine the feature maps resulting from the depthwise convolutions rather than linearly combining the kernels directly:
$$Y_i = \sum_{j=1}^{c_{in}} W_{i,j} \cdot K_j \ast X_j, \text{for } (i, j) \in [c_{out}] \times [c_{in}]$$

Although DS-CNNs are restricted-CNNs, but this restriction significantly reduces the parameter count while maintaining competitive performance.

Recent studies have revealed striking properties of depthwise convolutional filters in these networks. Trockman et al.~\cite{trockman2023understanding} observed that learned filters in their DS-CNN model ConvMixer exhibit highly structured covariance matrices. Furthermore, Babaiee et al.~\cite{babaiee2023unveiling} discovered that trained depthwise convolutional kernels across all layers of DS-CNNs converge into a few main clusters, each resembling the difference of Gaussian (DoG) functions and their first and second-order derivatives. The authors were able to classify the majority of the filters from state-of-the-art DS-CNN models. Building on this work, recent work.~\cite{babaiee2024master} introduced the "Master Key Filters Hypothesis," which proposes that depthwise filters in DS-CNNs exhibit generality across domains, architectures, and layer depths, challenging the conventional view that deeper layers become increasingly specialized.  Our work builds upon these findings by investigating the potential of using a limited number of unique filters in DS-CNNs, exploiting the observed clusterability in depthwise convolutional kernels, and moving towards finding the master key filters.

\textbf{Filter Diversity}. Filter pruning and compression techniques have been widely explored to reduce the computational complexity and memory footprint of CNNs~\cite{hoefler2021sparsity}. Structured pruning in CNNs is typically achieved by removing redundant filters~\cite{liebenwein2020provable}. These methods highlight the importance of "feature-map" diversity in CNNs, as removing redundant or less informative filters can lead to more efficient models without significant performance degradation. In contrast, our work is not attempting to reduce the number of parameters or computational cost, but rather investigating the role of "filter" diversity in DS-CNNs and challenging the assumption that a large number of unique filters is necessary for optimal performance. By discovering that a small set of carefully chosen filters can effectively replace a large number of learned filters in DS-CNNs, we are shedding light on the inherent limited diversity present in the learned depthwise filters.

\textbf{Scale-Space Theory.} Scale-space theory, which examines signals across different scales, was initially developed in the mid-1980s~\cite{koenderink1984structure} and has since become a fundamental concept in signal processing, particularly within the field of computer vision. Lindeberg~\cite{lindeberg2013scale} introduced a computational framework for visual receptive fields that exploits symmetry properties across space and time. This framework is compelling for two reasons~\cite{lindeberg2024approximation}: First, it offers a normative perspective on visual processing that closely aligns with the hierarchical stages observed in the visual systems of higher mammals\cite{lindberg2021}. Second, it provides a provable approach to capture the natural transformations of images over space and time~\cite{Lindeberg_2023}. Gaussian derivatives are the sole kernels satisfying isotropy (rotational invariance) and non-creativity (with respect to the causality principle) in scale-space theory~\cite{lindeberg2013scale}. Remarkably, our work reveals that the master key filters required for efficient performance in DS-CNNs consists of only 8 distinct filters: Gaussians, Difference of Gaussians (DoG, which can be approximated by the Laplacian of Gaussians), and first derivatives of Gaussians. This finding establishes a strong connection between the principles of scale-space theory and the design of CNN architectures.

\begin{figure*}[t]
  \centering
  \includegraphics[width=\linewidth]{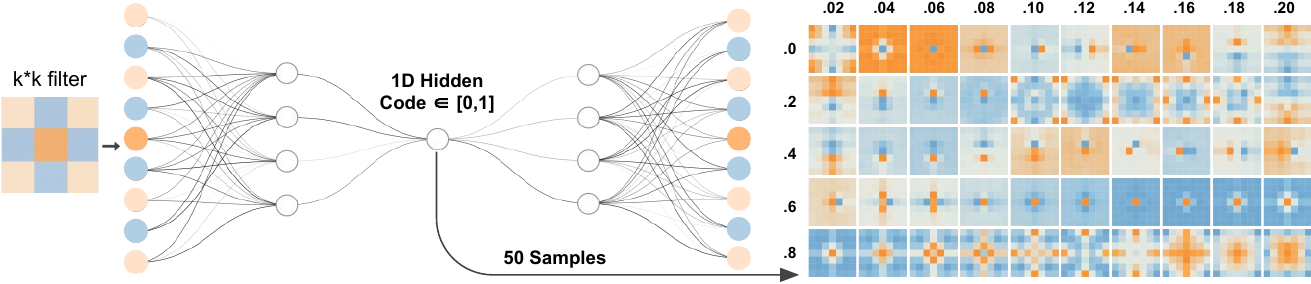}
   \caption{\textbf{Visualization of selecting candidate filters using autoencoder-based dimensionality reduction}. Left: An autoencoder compresses filters into a 1D hidden code. Right: Heatmap of 50 uniformly sampled candidate filters from 1D hidden code, generated by the decoder part of the autoencoder. These samples serve as the initial pool for our search for the master key filters.}
   \label{fig:process}
\end{figure*}

\section{Do We Need Thousands of Distinct Filters?}

In this section, we investigate whether employing thousands of unique filters is essential for maintaining the performance of DS-CNNs. In particular, we explore what is the impact on the performance of the network, when we replace the trained filters with a minimal set of distinct filter variations.

\subsection{The Quest for Master Key Filters}

In order to explore the possibility of reducing the number of distinct filters in DS-CNNs, we sought to distill the filters of trained models into a compact set. We collected filters from publicly available trained models of various sizes and employed an autoencoder to learn a compressed representation of these filters. The autoencoder was trained to encode each filter into a single dimension, following a similar procedure as the one described in \cite{babaiee2023unveiling}. This approach allowed us to capture the essential characteristics of the filters while significantly reducing their dimensionality.

To create a comprehensive filter set, we collected all the depthwise filters from every layer of the networks in our model bank. Each filter had a consistent size of $7\,{\times}\,7$. To ensure uniformity, we normalized the filters by first centering them and then by scaling their length to 1. This normalized filter set finally served as the training data for our autoencoder.

The autoencoder architecture comprises two primary components: an encoder and a decoder. The encoder consists of four intermediate layers, each followed by a leaky rectified linear unit (Leaky ReLU) activation function. These layers progressively compress and abstract the input filter representations. The final layer of the encoder, known as the code layer, employs a sigmoid activation function to map the compressed filter representations to values within the range [0, 1]. This mapping ensures that the encoded filters are bounded and compatible with the subsequent decoding process.

The decoder on the other hand, is responsible for reconstructing the original normalized filters from the encoded representations. It mirrors the structure of the encoder, with four intermediate layers that gradually upsample and expand the encoded features. The final layer of the decoder utilizes a hyperbolic tangent (tanh) activation function, which allows for accurate reconstruction of the normalized filters within [-1, 1]. This choice of activation function ensures that reconstructed filters maintain their centering and scale, aligning with the characteristics of the original normalized filters.

By training the autoencoder on the diverse set of normalized filters, we aim to learn a compact and meaningful representation of the filter space. The encoder captures the essential features and patterns present in the filters, while the decoder enables the reconstruction of filters from their encoded representations. This architecture facilitates the exploration of filter variations and the potential for reducing the number of distinct filters required in depthwise separable convolutional neural networks.


After training the autoencoder, we performed uniform sampling from the code layer. We took various numbers of samples, 50, 25, and 10, from the [0,1] interval to generate distinct filter sets. Using the decoder, we transformed these codes back into filter reconstructions.

A depthwise filter for the \( c \)-th channel can be denoted as ${F}_{c}$, where $c$ is the index of the channel. When flattened, \( {F}_{c} \) can be represented as a vector \(\mathbf{f}_{c}\,{\in}\,\mathbb{R}^{k^2}\), where $k\,{\times}\,k$ is the spatial dimension of the filter.
%
The matrix $F$ composed of these flattened vectors is $F = [\mathbf{f}_{1}, \mathbf{f}_{2},\ldots,\mathbf{f}_{C}]^T \in \mathbb{R}^{C \times k^2}$.

For each depthwise filter $F_{c}$ learned by the ConvNeXtV2-tiny model, we then conducted a linear approximation with respect to the decoded filters, by identifying the scalar coefficients $a$ and $b$, which minimized the Euclidean distance between the corresponding flattened filter vector $f_{c}$, and the linear combination $af_{c}' + b$, where $f_{c}'$ represents a flattened decoded-filter sample. The original filter was then substituted with the optimal linear combination $af_{c}' + b$ that exhibited the smallest distance to the original, thus preserving the filter's functional characteristics while reducing model complexity.

To solve for scalars \(a\) and \(b\) that minimize the distance between vectors \(f_c\) and $af_{c}' + b$, we use linear regression. Here, the goal is to determine the coefficients \(a\) and \(b\) for two vectors \(x\) and \(y\) such that by having $\tilde{y} = ax + b$ the length of the vector \(y - \tilde{y}\) is minimized. This problem has a well-known solution.

\begin{align} 
a = \frac{n \sum_{i=1}^n x_i y_i - \sum_{i=1}^n x_i \sum_{i=1}^n y_i}{n \sum_{i=1}^n x_i^2 - (\sum_{i=1}^n x_i)^2}
\label{eq:wiki1}
\quad
b = \frac{\sum_{i=1}^n y_i \sum_{i=1}^n x^2_i - \sum_{i=1}^n x_i \sum_{i=1}^n x_i y_i }{n \sum_{i=1}^n x^2_i - (\sum_{i=1}^n x_i)^2}
\end{align}

Calculating Equations \eqref{eq:wiki1} can be computationally intensive, especially when dealing with hundreds of thousands of filters. To reduce computational complexity, we can use a normalization trick. Since any linear shift of \(x\) does not alter the optimal \(\tilde{y}\), we normalize \(x\) using the transformation \(\hat{x} = \frac{x - \bar{x}}{||x - \bar{x}||}\). With this normalization, \(\sum_{i=1}^n \hat{x}_i = 0\) and \(\sum_{i=1}^n \hat{x}_i^2 = 1\), allowing us to simplify Equation\eqref{eq:wiki1}.

\begin{align} 
a = \frac{n \sum_{i=1}^n x_i y_i}{n}
\label{eq:simple1} 
= \langle x, y \rangle \quad
b = \frac{\sum_{i=1}^n y_i}{n}
= \bar{y}
\end{align}

Consequentially, Given the vectors \(\hat{x}_1, \hat{x}_2, \ldots, \hat{x}_n\) as the rows of matrix \(\hat{X}\) and the vectors \(y_1, y_2, \ldots, y_m\) as the columns of matrix \(Y\), we introduce the vector \(y_{\text{mean}}\), which contains the means \(\Bar{y}_1, \Bar{y}_2, \ldots, \Bar{y}_m\). Using these, we can calculate the coefficients \(a_{ij}\) and \(b_{ij}\) for each pair of \(x_i\) and \(y_j\) through matrix multiplication.

\begin{align} 
A = \hat{X}Y \quad
B = y_{\text{mean}}\mathbf{1}^\top
\end{align}
For each layer, with the set of depthwise filter vectors matrix $F$ and the sample filter vectors matrix $F'$, we calculate the coefficients as above to find the closest linear approximation.

We chose linear shifts for approximating the original filters because they preserve the heatmap and essential characteristics of the filters. By applying a linear shift to a filter sample, we maintain the spatial structure and relative importance of different regions within the filter.

In order to evaluate the impact of replacing the original filters with their linear approximations derived from the sampled filter set, we assessed the performance of the modified models on the test set. In Table~\ref{tab:models_acc} we present the accuracy of models with varying sizes from the ConvNeXtv2 and Hornet~\cite{hornet} families, along with their accuracy after their filter replacement. Quite remarkably, when replacing all the filters of the models with approximations based on only 50 sampled filters, the model performance remains robust, even without any fine-tuning. This resilience is particularly evident for larger model sizes. In the case of ConvNeXtv2 Huge, replacing nearly 50K filters with just 50 sampled filters results in less than a two percent accuracy drop, without any fine-tuning.


As expected, reducing the number of sampled filters leads to a larger accuracy gap, and the ConvNeXtv2 models struggle to perform well when using an small set of only 10 filters. However, it is important to note that the filter samples used in this experiment were obtained through uniform sampling from the code layer of the autoencoder. This immediately raises the following question: \textit{Is there a more strategically selected set of filter samples which can yield a better performance?}

To elucidate this question, we focused on the ConvNeXt-v2-Tiny model and conducted a greedy search on a set of 50 filter samples. We began with the 50 uniformly sampled filters and iteratively removed filters one by one. Figure~\ref{fig:process} illustrates the 50 filter samples used in this search. At each iteration, we evaluated the model accuracy after removing each individual filter and eliminated the one whose removal resulted in the smallest accuracy drop. This process was repeated for all filters.

\begin{wrapfigure}{r}{0.5\linewidth}
    \centering
    \includegraphics[width=\linewidth]{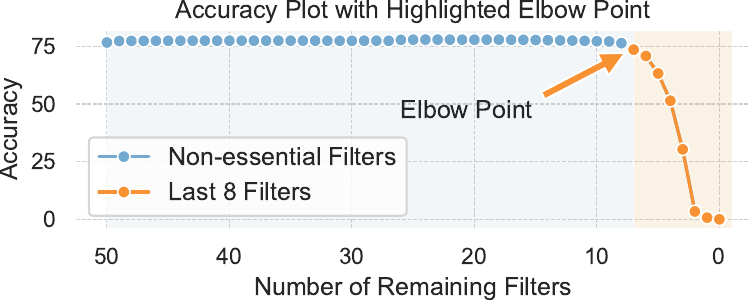}
    \caption{\textbf{Our systematic greedy search for the essential filters}. While the removal of most filters did not noticeably change accuracy, 8 of them were essential, consistently in all models we tested.}
    \label{fig:acc_curve}
    \vspace{-20pt}
\end{wrapfigure}

The accuracy plot during the greedy search, as shown in Figure~\ref{fig:acc_curve}, reveals an interesting trend. The model's accuracy remains relatively stable until the last 10 filters are removed, with the curve exhibiting a distinct elbow around 8 samples. This observation suggests that a small subset of only 10 filters is playing a crucial role in maintaining the performance of the model.
To further refine the search, we selected the 10 best-performing filters from the previous step and expanded our search space by sampling 4 additional filters around each of these 10 filters. This local exploration allows us to fine-tune the selection of filters and capture any potential variations that may enhance performance.

\begin{figure*}[t]
\centering
\includegraphics[width=1\linewidth]{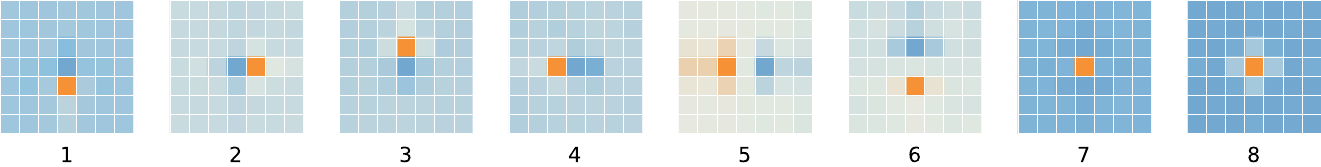}
\caption{\textbf{Heatmap visualization of the eight universal filters discovered through systematic greedy search on the ConvNeXtv2 tiny model.} Our empirical analysis demonstrates that DS-CNN filters predominantly converge to linear shifts (ax+b) of one of these eight filters, regardless of architecture or dataset. Filters 1-4 display central difference operator characteristics, and filters 5-8 correspond to established mathematical image processing fundamentals (See Figure \ref{fig:functions}).}
\label{fig:8filters} 

\end{figure*}

We then conducted a second round of the greedy search using this expanded set of filters. The search converged to a set of 8 filters located just after the curve elbow. These 8 filters, as shown in Figure~\ref{fig:8filters}, represent a highly informative subset that can effectively replace the original large set of filters, while minimally impacting the model's accuracy.


The last row of Table~\ref{tab:models_acc} showcases the accuracy of the models, when their filters are replaced by the 8 filter transformations, obtained from the greedy search. Remarkably, the results demonstrate that the  ConvNeXtv2 models accuracy achieved with these 8 filters, surpasses even the performance of the 25-filters-sample set. This finding underscores the effectiveness of the greedy search approach in identifying a highly discriminative subset of filters. Moreover, it highlights the potential for replacing a large number of filters, up to 50K in the case of ConvNeXt v2 models, with just 8 strategically selected filters, while maintaining an acceptable performance.


\begin{table}[t]
  \caption{\textbf{Performance comparison when thousands of trained filters are replaced with linear shifts (ax+b) from candidate filters.} \emph{Without \textbf{any} fine-tuning} (For trained model see Table \ref{tab:maintable}), models with just 8 selected filters from our greedy search on \emph{ConvNeXtv2 Tiny} maintain remarkably high accuracy (e.g., only 3.5\% drop for ConvNeXtv2 Huge despite reducing from 50k to 8 unique filter patterns) and even on different architecture, HorNet. Concidering models' high sensitivity to filter alterations, this is evidence that DS-CNN filters predominantly converge to these filters.}
\vspace*{2mm}
  \label{tab:models_acc}
  \setlength{\tabcolsep}{7pt} 
  \renewcommand{\arraystretch}{1.3}
  \centering
  \begin{tabular}{@{}l|lllll|ll@{}}
  & \multicolumn{5}{c|}{ConvNeXt} & \multicolumn{2}{c}{Hornet} \\
    \textbf{ConvNeXtv2 Models} & \textbf{Pico} & \textbf{Tiny} & \textbf{Base} & \textbf{Large} & \textbf{Huge} & \textbf{Tiny} & \textbf{Small} \\
    Number of Filters & 2\,944 & 6\,624 & 18\,048 & 27\,072 & 49\,632 & 11\,488 & 17\,232 \\
    \midrule
    \rowcolor{myOrange!17}
    Original Acc & 80.3\% & 83.0\% & 84.9\% & 85.8\% & 86.3\% & 82.3\% & 83.5\%\\
    Acc with 50 candidates& 75.0\% & 75.4\% & 80.5\% & 83.2\% & 84.0\% & 79.4\% & 81.3\% \\
    Acc with 25 candidates& 72.0\% & 66.9\% & 72.8\% & 79.6\% & 80.4\% & 78.3\% & 80.9\%\\
    Acc with 10 candidates& 23.4\% & 1.0\% & 1.4\% & 3.0\% & 2.0\% & 66.3\% & 70.5\%\\
    \rowcolor{myOrange!17}
    Acc with 8 (greedy search) & 73.1\% &76.7\% & 79.3\% & 81.2\% & 82.8\% & 76.0\% & 78.1\%\\
    Acc with 8 random filters & 0.11\% & 0.10\% & 0.10\% & 0.12\% & 0.09\%  & 0.96\% & 1.0\% \\
  \end{tabular}
\vspace{-3ex}
\end{table}

To validate the generalizability of these 8 filter samples, we extended our experiments to the ConvNeXt-v2-Pico model, which represents a different model size. In this case also, we arrived at a similar set of 8 filter samples, indicating the robustness and transferability of our findings across different model architectures. The consistency of the 8 filter samples across different model sizes suggests that these filters capture fundamental and generalizable patterns in the data. It hints at the existence of a set of universal filters that can effectively represent the essential information required for accurate classification.

The remarkable performance maintained when replacing thousands of trained filters with just our 8 master key filters cannot be coincidental. Noting models' high sensitivity to filter alterations (evidenced by the catastrophic performance drop with 8 random filters), strongly indicates that DS-CNN filters predominantly converge to fundamental patterns during training. Even more compelling, these 8 filters (discovered exclusively from ConvNextV2 Tiny) transfer seamlessly to architecturally distinct models like Hornet. This fascinating cross-architecture generalization suggests that DS-CNNs naturally gravitate toward a mathematically well-defined universal filter code that captures fundamental visual processing operations.

\begin{figure*}[t]
  \centering
  \begin{subfigure}{0.22\linewidth}
    \includegraphics[width=\linewidth]{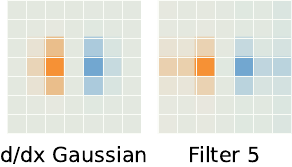}
  \end{subfigure}
  \hfill
  \begin{subfigure}{0.22\linewidth}
    \includegraphics[width=\linewidth]{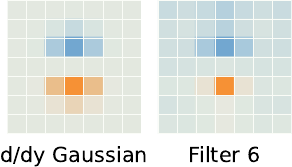}
  \end{subfigure}
  \hfill
  \begin{subfigure}{0.22\linewidth}
    \includegraphics[width=\linewidth]{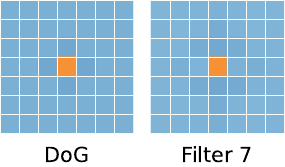}
  \end{subfigure}
  \hfill
  \begin{subfigure}{0.22\linewidth}
    \includegraphics[width=\linewidth]{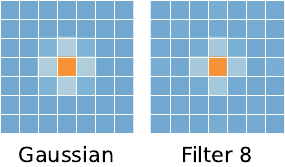}
  \end{subfigure}
  \caption{\textbf{Correspondence between the empirically discovered filters (5-8) and their theoretical mathematical counterparts.} Left column shows the idealized mathematical forms: first derivatives of Gaussians in x and y directions (filters 5-6), Difference of Gaussians (filter 7), and Gaussian function (filter 8). Right column shows our discovered filters that closely approximate these operators, demonstrating the network's natural convergence toward established visual processing primitives.}
\label{fig:functions}
\end{figure*}


\subsection{Understanding the Eight Filters}

\begin{table}[t]
  \caption{\textbf{ImageNet Top-1 accuracy comparison between conventional trainings and our 8-filter constraint.} Models restricted to using only our 8 unique filters (plus learnable bias terms) achieve comparable accuracy to their fully-trained counterparts. The consistent performance across different architectures (ConvNeXtv2 and Hornet) demonstrates the universality of these fundamental filters.}
\label{tab:maintable}
  \centering
  \begin{threeparttable}
  \setlength{\tabcolsep}{12pt} 
  \renewcommand{\arraystretch}{1.3}

 \begin{tabular}{@{}l|llll|l@{}}
  & \multicolumn{4}{c|}{ConvNeXtv2} & \multicolumn{1}{c}{Hornet} \\
    \textbf{Models} & \textbf{Pico} & \textbf{Tiny} & \textbf{Base} & \textbf{Large} & \textbf{Tiny} \\
    Number of Original Filters & 2\,944 & 6\,624 & 18\,048 & 27\,072 & 11\,488 \\
    \midrule
    Original model with FCMAE\tnote{1}   & 80.3\% & 82.9\% & 84.9\% & 85.8\% & --- \\
    Original model              & 79.7\% & 82.5\% & 84.3\% & 84.5\% & 82.3\%\\
    \rowcolor{myOrange!17} Our 8 unique filters + bias& 80.2\% & 82.7\% & 84.6\% & 85.4\% & 81.8\% \\
  \end{tabular}
  \begin{tablenotes}
\item[1] FCMAE {\footnotesize(Fully Convolutional Masked Autoencoder Framework)} is a heavy pretraining.
\end{tablenotes}
\end{threeparttable}
\end{table}

This subsection investigates into the functional characteristics of the eight filters identified through our systematic greedy search, and which have proven to be very effective across various datasets. By analyzing these filters, we aim to understand their resemblance to traditional image-processing operators and their potential roles in effective feature extraction within the network.

\textbf{Filters 1-4:} These 4 filters in Figure exhibit characteristics reminiscent of central difference operators, commonly used for approximating Gaussian derivatives discretely. The arrangement and weights of these filters mimic the theoretical models used in edge detection and texture analysis.
    
\textbf{Filters 5-6:} These 2 filters strongly resemble 1st order Gaussian derivatives along the $x$ and $y$ axis. These filter contribute more pronounced spatial smoothing than previous filters. This characteristic enables these filters to capture broader and more varied textural information from the input images, potentially allowing for a better generalization across different visual contexts.

\textbf{Filters 7:} This filter resemble a 2-D discrete analogue of the Difference of Gaussians (DoG) due to its positive center with slightly negative surround. The DoG filter, often approximated by the Laplacian of the Gaussian in digital image processing, is crucial for blob detection and bar pattern recognition in images. These filter likely contribute to the model's ability to differentiate areas of rapid intensity change, enhancing edge and contour detection.
    
\textbf{Filter 8:} This filter closely aligns with a very fine-scaled Gaussian kernel. In image processing, Gaussian kernels are smoothing filters used to reduce noise and detail. This results in a blurred image that preserves edges better than uniform filters. Gaussian filters are mathematically proven to be the only function for scale-space representation. 

\textbf{Filters formal definition:}
For completeness, we provide below the formal definition of the continuous functions corresponding to the 2D Gaussian, the 2D derivative of the Gaussian along the $x$ and the $y$ axis, respectively, and the 2D difference of Gaussians (DoG, Laplacian, Mexican hat):
\[
\begin{array}{cllcll}
    \text{Gaussian:}         &G(x, y)&=&e^{-(x^2 + y^2)/2\sigma^2}&\quad \text{$\Delta$Gaussian:} &\text{DoG}(x, y)=G_1(x, y)-G_2(x, y)
\end{array}
\]


These formulations were used to construct the last four filters as depicted in Figure \ref{fig:functions}, with the exception of the DoG, for which its approximation, the Ricker wavelet, was used for simplicity. The reconstructed filters bear a strong resemblance to those discovered through our encoding and greedy search methods, validating our hypothesise. in function approximation and emphasizing the practical relevance of traditional image processing theories in modern deep learning architectures. 

\textbf{Functional Approximation and Construction:} We reconstructed the lower four filters using theoretical formulas typically associated with these image processing techniques, depicted in Figure~\ref{fig:functions}. The reconstructed filters bear a strong resemblance to those discovered through our encoding and greedy search methods, validating our approach in filter selection and emphasizing the practical relevance of traditional image processing theories in modern deep learning architectures. 



\section{Experiments}

So far, we've identified 8 unique filters that, when used as linear approximations to replace the filters of trained models, maintain relatively stable accuracy despite this dramatic change in model parameters. These findings naturally lead to a key question: \textit{Can models be successfully trained from initialization with just these 8 filter types kept frozen throughout training?} In this section, we present experimental evaluations on ImageNet and additional datasets to investigate this question.

\subsection{ImageNet}

The results in Table~\ref{tab:models_acc} demonstrate that model accuracy remains stable despite significantly reduced filter diversity. In these experiments, model filters are linear shifts of one of the 8 identified filters, mathematically expressed as $a(x+b)$. Given the architecture of DS-CNNs, the coefficient 
$a$ can be transferred to the fully-connected layers following depthwise convolutions (the following pointwise layer), effectively simplifying the filters to $x+b$, where $b$ acts as a learnable bias. This insight motivated us to train models from scratch using only these 8 fixed filters with learnable biases.

\textbf{Training with Only 8 Frozen Filters}

To investigate the effectiveness of our 8 candidate filters, we trained ConvNeXtv2 models from scratch, initializing each layer's filters with these 8 filters while allowing only the bias terms to be trainable. We followed the same 300-epoch training pipeline described in the original paper~\cite{ConvNeXtV2}, with the critical difference that all convolutional filters remained frozen throughout training.
Table~\ref{tab:maintable} presents our results. Remarkably, the ConvNeXtv2 Tiny model with only 8 types of filters achieved an accuracy of 82.7\%, merely 0.2\% lower than the model trained with 6,624 trainable filters and FCMAE pretraining. Similarly, the smaller ConvNeXtv2 Pico model with 8 types of frozen filters reached 80.2\% accuracy, just 0.1\% below the model with 2,944 trainable filters.

To validate the generalizability of our findings across different architectures, we conducted an additional experiment with the Hornet model~\cite{hornet}—a DS-CNN with substantially different structure than the ConvNext family. The Hornet Tiny model with only our 8 filters achieved 81.8\% accuracy compared to 82.3\% for the original model, representing only a 0.5\% drop. Notably, these 8 filters were derived exclusively from ConvNext models through greedy search on the ConvNeXtv2 Tiny model, yet transferred effectively to Hornet without modification.

It is worth emphasizing that despite our significant architectural modification, these experiments used the original training hyperparameters for each model. A dedicated hyperparameter search optimized for this fixed-filter approach could potentially enhance results further.

\begin{table}[!htbp]
  \caption{\textbf{Cross-dataset evaluation} demonstrating the superiority of our universal filter approach on smaller datasets. The ConvNeXt Femto model restricted to using only 8 unique frozen filters outperforms both models trained from scratch and those using transferred ImageNet filters, on Oxford Pets and Oxford Flowers.  We evaluate multiple model sizes (Atto, Femto, Pico, Tiny) for Flowers and Pets datasets to verify that our observed advantage is consistent across architectures of varying capacity and not merely an artifact of dataset size limitations.}
  \label{tab:small_datasets}
  \setlength{\tabcolsep}{2.5pt} 
  \renewcommand{\arraystretch}{1.1}
  \centering
  \begin{tabular}{@{}l|l|l|llll|llll@{}}
   \textbf{Dataset}  & \multicolumn{1}{c|}{{CIFAR10}}  & \multicolumn{1}{c|}{{STL-10}} & \multicolumn{4}{c|}{{Oxford Flowers}} & \multicolumn{4}{c}{{Oxford Pets}}  \\
   \small{\# Training Set Size}  & \multicolumn{1}
   {c|} {\small{50000}}  & \multicolumn{1}{c|} {\small{5000}} & \multicolumn{4}{c|} {\small{2040}} & \multicolumn{4}{c} {\small{3680}}  \\
   \hline
   \diagbox[width=12em]{Filters}{Model}     & 
   Femto  & Femto  & Atto & Femto & Pico & Tiny & Atto & Femto & Pico & Tiny \\
  
       \hline
        Original (normal traning) &96.9&80.4&63.3&66.0&60.2&75.7    &38.4&36.3&40.1&65.4 \\
        ImageNet Transferring &\textbf{97.1}&\textbf{83.2}& 72.2&73.2&74.8&81.8    &58.5&56.0&66.3&80.1\\
        \rowcolor{myOrange!17}Our 8 Unique Filters
                &96.3&83.1&\textbf{77.8}&\textbf{77.7}&\textbf{77.2}&\textbf{85.1}   &\textbf{66.5}&\textbf{66.4}&\textbf{72.8}&\textbf{81.8}\\

  \end{tabular}
\end{table}

\subsection{Other Datasets}
To investigate the generalizability of our findings, we extend our experiments to other datasets and compare the performance of the ConvNeXt Femto across various settings.

\textbf{Datasets and Settings.} We evaluate the low filter variety on four datasets: CIFAR-10~\cite{cifar10}, Flowers~\cite{flowers}, Pets~\cite{pets}, and STL-10~\cite{stl10}. These datasets have smaller scales compared to ImageNet, with the size of training sets ranging from 2040 to 50000 samples. We use the ConvNeXt Femto model as our base architecture for all datasets, and additionally use ConvNeXt Atto, Pico, and Tiny for the Flowers and Pets datasets. For a fair comparison, we train the model on all datasets for 300 epochs, following the training parameters from the ConvNeXt paper~\cite{convnext}, and keep the training settings consistent across all runs and datasets. For each dataset, we first train the model to obtain baseline accuracy. We then evaluate two filter initialization strategies: (1) depthwise filters transferred from a ConvNeXt model pretrained on ImageNet, and (2) eight frozen filter types trained from scratch. Table~\ref{tab:small_datasets} shows the resulting accuracies for each setting.

\textbf{Results.} The results demonstrate the effectiveness of using the 8 frozen filters across different datasets. Notably, the performance improvement becomes more pronounced as the dataset size decreases. For the Flowers and Pets datasets, the frozen 8 filter types achieve remarkable improvements of up to 11\% and 34.5\%, respectively, compared to the baseline model. Interestingly, on these smaller datasets, the eight frozen filter types even outperform the transferred filters from the model trained on ImageNet. To further evaluate and verify the performance of our 8 filters on these datasets, we used other sizes of the ConvNeXt model, the results of which showed consistent superior performance on all sizes.

This observation suggests that the carefully selected filter types capture fundamental patterns that are highly relevant to the task at hand, even when the dataset size is limited. This finding has significant implications for scenarios where training data is scarce or computational resources are limited.


\section{Conclusions}
This paper extends the "Master Key Filters Hypothesis" by identifying a set of just 8 filters. While conventional DS-CNNs employ thousands of distinct trained filters, our analysis reveals these filters predominantly converge to linear shifts (ax+b) of one of the filters in our discovered set. This finding significantly narrows the theoretical space proposed in previous work. The discovered filters closely match established mathematical forms: Difference of Gaussians, Gaussians, and their derivatives, creating a bridge between classical computer vision theory and modern deep learning practice. This correspondence to structures found in both scale-space theory and mammalian visual systems suggests DS-CNNs inherently rediscover optimal operators aligned with natural image statistics.

Our systematic experiments demonstrate that networks initialized with these 8 frozen filters achieve over 80\% ImageNet accuracy. Particularly noteworthy is the superior performance of our filters on smaller datasets, where models initialized with our filters outperform even ImageNet transfer learning. This suggests that these filters encode fundamental visual processing primitives that transcend specific datasets and visual domains, offering a novel approach to transfer learning.


\textbf{Future Work} may explore direct optimization approaches to refine our master key filter set. While our 8 filters demonstrate impressive performance, systematic optimization might further improve their effectiveness or reduce their number. The observed constraint on filter diversity invites us to rethink the fundamental principles governing these architectures, potentially leading to insights about the complementary roles of depthwise spatial filters and pointwise channel mixing layers and opening opportunities for novel architecture designs.


{
\small

\bibliography{references}
\bibliographystyle{plain}
}

\newpage
\appendix

\section{Technical Appendices and Supplementary Material}

\subsection{Experimental Settings}

\begin{table}[h]
\caption{Training (t) and fine-tuning (ft) hyperparameters used in Section 4.2 experiments for ConvNeXtv2 Tiny model, taken from~\cite{ConvNeXtV2}.}
\centering
\begin{tabular}{l|l}
\textbf{config} & \textbf{value} \\
\midrule
optimizer & AdamW \\
base learning rate & 8e-4 \\
weight decay & 0.05 \\
optimizer momentum & $\beta_1, \beta_2=0.9, 0.999$ \\
layer-wise lr decay \cite{clark2020electra,bao2022beit} & 0.9 \\
batch size & 1024 \\
learning rate schedule & cosine decay \\
warmup epochs & (t) 40, (ft) 3\\
training epochs & (t) 300, (ft) 100 \\
augmentation & RandAug (9, 0.5) \cite{cubuk2020randaugment} \\
label smoothing \cite{szegedy2016rethinking} & 0.1 \\
mixup \cite{zhang2018mixup} & 0.8 \\
cutmix \cite{yun2019cutmix} & 1.0 \\
drop path \cite{larsson2017fractalnet} & 0.2 \\
head init \cite{touvron2021training} & 0.001 \\
ema & 0.9999 \\
\end{tabular}

\label{tab:finetune_afpn}
\end{table}

\begin{table}[h]
\caption{Training (t) and fine-tuning (ft) hyperparameters used in Section 4.2 experiments for ConvNeXtv2 Pico model, taken from~\cite{ConvNeXtV2}.}

\centering
\begin{tabular}{l|l}
\textbf{config} & \textbf{value} \\
\midrule
optimizer & AdamW \\
base learning rate & 2e-4 \\
weight decay & 0.3 \\
optimizer momentum & $\beta_1, \beta_2=0.9, 0.999$ \\
layer-wise lr decay \cite{clark2020electra,bao2022beit} & 0.9 \\
batch size & 1024 \\
learning rate schedule & cosine decay \\
warmup epochs & 0 \\
training epochs & (t) 600, (ft) 100 \\
augmentation & RandAug (9, 0.5) \cite{cubuk2020randaugment} \\
label smoothing \cite{szegedy2016rethinking} & 0.2 \\
mixup \cite{zhang2018mixup} & 0.3 \\
cutmix \cite{yun2019cutmix} & 0.3 \\
drop path \cite{larsson2017fractalnet} &  0.0  \\
head init \cite{touvron2021training} & 0.001 \\
ema & 0.9999 \\
\end{tabular}
\label{tab:finetune_tiny}
\end{table}

\begin{table}[h]
  \caption{Information of Datasets used in the study and sample sizes, in training set size descending order.}
  \label{tab:datasets}
  \setlength{\tabcolsep}{5pt} 
  \centering
  \begin{tabular}{@{}l|lll@{}}
    \textbf{Dataset} & \textbf{Classes} & \textbf{Train Samples} & \textbf{Test Samples} \\
    \midrule
    ImageNet & 1000 & 1.2 million & 50,000 \\
    CIFAR-10~\cite{cifar10} & 10 & 50,000 & 10,000  \\
    STL-10~\cite{stl10} & 10 & 5,000 & 8,000 \\
    Oxford-IIIT Pets~\cite{pets} & 37 & 3,680 & 3,369  \\
    Oxford 102 Flowers~\cite{flowers} & 102 & 2,040 & 6,149 \\
  \end{tabular}
\end{table}

\begin{table}[h]
\caption{Training hyperparameters used in Section 4.2 experiments. The setting is taken from ConvNeXt \cite{convnext}.}

\centering
\begin{tabular}{l|l}
\textbf{config} & \textbf{value} \\
\midrule
optimizer & AdamW \\
base learning rate & 4e-3 \\
weight decay & 0.05 \\
optimizer momentum & $\beta_1, \beta_2=0.9, 0.999$ \\
batch size & 4096 \\
training epochs & 300 \\
learning rate schedule & cosine decay \\
warmup epochs & 50 \\
warmup schedule & linear \\
layer-wise lr decay & None \\
randaugment  & (9, 0.5) \\
mixup  & 0.8 \\
cutmix  & 1.0 \\
random erasing & 0.25 \\
label smoothing  & 0.1 \\
layer scale  & 1e-6 \\
head init scale  & None \\
gradient clip & None \\
\end{tabular}

\label{tab:training_recipe}
\end{table}

\subsection{8 Master Key Filters}

In Figure~\ref{fig:8filters} we provide the full numerical values of the 8 discovered universal filters, each of size $7 \times 7$. These filters were derived from a greedy search over encoded depthwise filters, as described in Section 3.1, and used in all experimental evaluations (Sections 4.1 and 4.2).

\begin{figure}
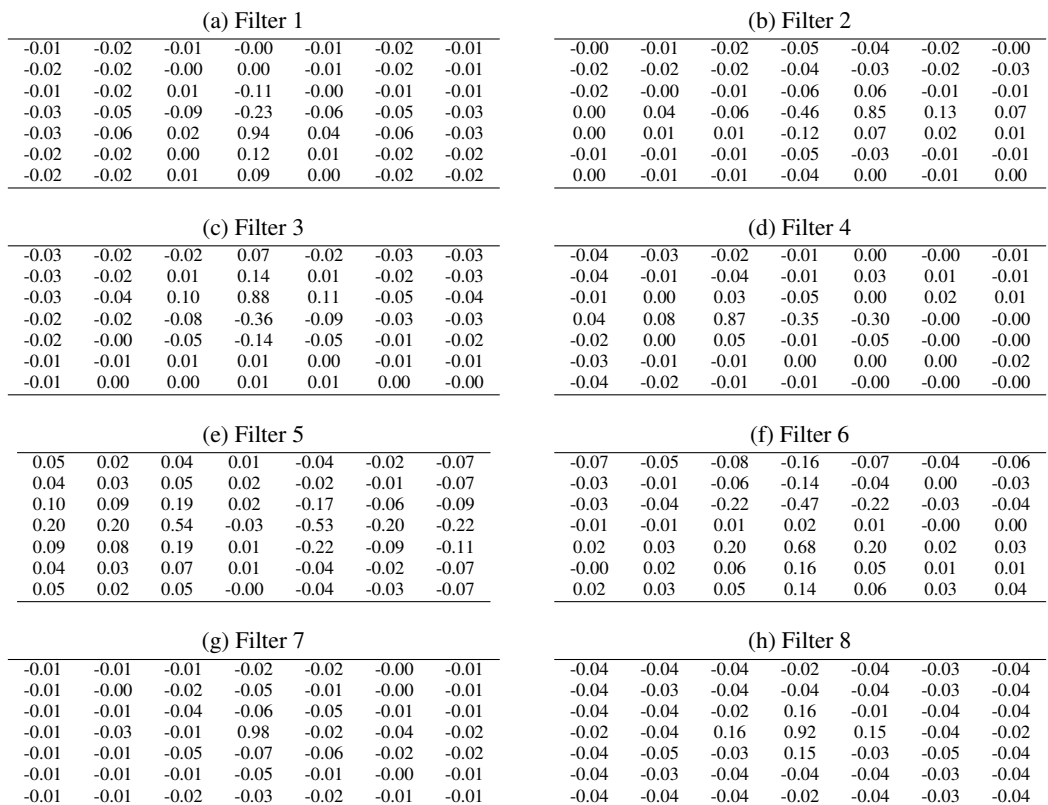

    \centering
    \scriptsize
    
    \begin{subfigure}[b]{0.48\textwidth}
        \centering
        \caption{Filter 1}
        \begin{tabular}{ccccccc}
            \hline
            -0.01 & -0.02 & -0.01 & -0.00 & -0.01 & -0.02 & -0.01 \\
            -0.02 & -0.02 & -0.00 & 0.00 & -0.01 & -0.02 & -0.01 \\
            -0.01 & -0.02 & 0.01 & -0.11 & -0.00 & -0.01 & -0.01 \\
            -0.03 & -0.05 & -0.09 & -0.23 & -0.06 & -0.05 & -0.03 \\
            -0.03 & -0.06 & 0.02 & {0.94} & 0.04 & -0.06 & -0.03 \\
            -0.02 & -0.02 & 0.00 & 0.12 & 0.01 & -0.02 & -0.02 \\
            -0.02 & -0.02 & 0.01 & 0.09 & 0.00 & -0.02 & -0.02 \\
            \hline
        \end{tabular}
    \end{subfigure}
    \hfill
    \begin{subfigure}[b]{0.48\textwidth}
        \centering
        \caption{Filter 2}
        \begin{tabular}{ccccccc}
            \hline
            -0.00 & -0.01 & -0.02 & -0.05 & -0.04 & -0.02 & -0.00 \\
            -0.02 & -0.02 & -0.02 & -0.04 & -0.03 & -0.02 & -0.03 \\
            -0.02 & -0.00 & -0.01 & -0.06 & 0.06 & -0.01 & -0.01 \\
            0.00 & 0.04 & -0.06 & -0.46 &  {0.85} & 0.13 & 0.07 \\
            0.00 & 0.01 & 0.01 & -0.12 & 0.07 & 0.02 & 0.01 \\
            -0.01 & -0.01 & -0.01 & -0.05 & -0.03 & -0.01 & -0.01 \\
            0.00 & -0.01 & -0.01 & -0.04 & 0.00 & -0.01 & 0.00 \\
            \hline
        \end{tabular}
    \end{subfigure}
    
    \vspace{0.5em}
    \begin{subfigure}[b]{0.48\textwidth}
        \centering
        \caption{Filter 3}
        \begin{tabular}{ccccccc}
            \hline
            -0.03 & -0.02 & -0.02 & 0.07 & -0.02 & -0.03 & -0.03 \\
            -0.03 & -0.02 & 0.01 & 0.14 & 0.01 & -0.02 & -0.03 \\
            -0.03 & -0.04 & 0.10 &  {0.88} & 0.11 & -0.05 & -0.04 \\
            -0.02 & -0.02 & -0.08 & -0.36 & -0.09 & -0.03 & -0.03 \\
            -0.02 & -0.00 & -0.05 & -0.14 & -0.05 & -0.01 & -0.02 \\
            -0.01 & -0.01 & 0.01 & 0.01 & 0.00 & -0.01 & -0.01 \\
            -0.01 & 0.00 & 0.00 & 0.01 & 0.01 & 0.00 & -0.00 \\
            \hline
        \end{tabular}
    \end{subfigure}
    \hfill
    \begin{subfigure}[b]{0.48\textwidth}
        \centering
        \caption{Filter 4}
        \begin{tabular}{ccccccc}
            \hline
            -0.04 & -0.03 & -0.02 & -0.01 & 0.00 & -0.00 & -0.01 \\
            -0.04 & -0.01 & -0.04 & -0.01 & 0.03 & 0.01 & -0.01 \\
            -0.01 & 0.00 & 0.03 & -0.05 & 0.00 & 0.02 & 0.01 \\
            0.04 & 0.08 &  {0.87} & -0.35 & -0.30 & -0.00 & -0.00 \\
            -0.02 & 0.00 & 0.05 & -0.01 & -0.05 & -0.00 & -0.00 \\
            -0.03 & -0.01 & -0.01 & 0.00 & 0.00 & 0.00 & -0.02 \\
            -0.04 & -0.02 & -0.01 & -0.01 & -0.00 & -0.00 & -0.00 \\
            \hline
        \end{tabular}
    \end{subfigure}
    
    \vspace{0.5em}
    \begin{subfigure}[b]{0.48\textwidth}
        \centering
        \caption{Filter 5}
        \begin{tabular}{ccccccc}
            \hline
            0.05 & 0.02 & 0.04 & 0.01 & -0.04 & -0.02 & -0.07 \\
            0.04 & 0.03 & 0.05 & 0.02 & -0.02 & -0.01 & -0.07 \\
            0.10 & 0.09 & 0.19 & 0.02 & -0.17 & -0.06 & -0.09 \\
            0.20 & 0.20 &  {0.54} & -0.03 & -0.53 & -0.20 & -0.22 \\
            0.09 & 0.08 & 0.19 & 0.01 & -0.22 & -0.09 & -0.11 \\
            0.04 & 0.03 & 0.07 & 0.01 & -0.04 & -0.02 & -0.07 \\
            0.05 & 0.02 & 0.05 & -0.00 & -0.04 & -0.03 & -0.07 \\
            \hline
        \end{tabular}
    \end{subfigure}
    \hfill
    \begin{subfigure}[b]{0.48\textwidth}
        \centering
        \caption{Filter 6}
        \begin{tabular}{ccccccc}
            \hline
            -0.07 & -0.05 & -0.08 & -0.16 & -0.07 & -0.04 & -0.06 \\
            -0.03 & -0.01 & -0.06 & -0.14 & -0.04 & 0.00 & -0.03 \\
            -0.03 & -0.04 & -0.22 & -0.47 & -0.22 & -0.03 & -0.04 \\
            -0.01 & -0.01 & 0.01 & 0.02 & 0.01 & -0.00 & 0.00 \\
            0.02 & 0.03 & 0.20 &  {0.68} & 0.20 & 0.02 & 0.03 \\
            -0.00 & 0.02 & 0.06 & 0.16 & 0.05 & 0.01 & 0.01 \\
            0.02 & 0.03 & 0.05 & 0.14 & 0.06 & 0.03 & 0.04 \\
            \hline
        \end{tabular}
    \end{subfigure}
    
    \vspace{0.5em}
    \begin{subfigure}[b]{0.48\textwidth}
        \centering
        \caption{Filter 7}
        \begin{tabular}{ccccccc}
            \hline
            -0.01 & -0.01 & -0.01 & -0.02 & -0.02 & -0.00 & -0.01 \\
            -0.01 & -0.00 & -0.02 & -0.05 & -0.01 & -0.00 & -0.01 \\
            -0.01 & -0.01 & -0.04 & -0.06 & -0.05 & -0.01 & -0.01 \\
            -0.01 & -0.03 & -0.01 &  {0.98} & -0.02 & -0.04 & -0.02 \\
            -0.01 & -0.01 & -0.05 & -0.07 & -0.06 & -0.02 & -0.02 \\
            -0.01 & -0.01 & -0.01 & -0.05 & -0.01 & -0.00 & -0.01 \\
            -0.01 & -0.01 & -0.02 & -0.03 & -0.02 & -0.01 & -0.01 \\
            \hline
        \end{tabular}
    \end{subfigure}
    \hfill
    \begin{subfigure}[b]{0.48\textwidth}
        \centering
        \caption{Filter 8}
        \begin{tabular}{ccccccc}
            \hline
            -0.04 & -0.04 & -0.04 & -0.02 & -0.04 & -0.03 & -0.04 \\
            -0.04 & -0.03 & -0.04 & -0.04 & -0.04 & -0.03 & -0.04 \\
            -0.04 & -0.04 & -0.02 & 0.16 & -0.01 & -0.04 & -0.04 \\
            -0.02 & -0.04 & 0.16 &  {0.92} & 0.15 & -0.04 & -0.02 \\
            -0.04 & -0.05 & -0.03 & 0.15 & -0.03 & -0.05 & -0.04 \\
            -0.04 & -0.03 & -0.04 & -0.04 & -0.04 & -0.03 & -0.04 \\
            -0.04 & -0.04 & -0.04 & -0.02 & -0.04 & -0.03 & -0.04 \\
            \hline
        \end{tabular}
    \end{subfigure}
    
    \caption{The 8 filters.}
    \label{fig:filters}
\end{figure}

\subsection{Experimental Compute Resources}
We used 2 NVIDIA TITAN RTX GPUs for experiments on other datasets. For ImageNet training and fine-tuning we used 8 NVIDIA TITAN RTX GPUs.


\clearpage
\section*{NeurIPS Paper Checklist}

\begin{enumerate}

\item {\bf Claims}
    \item[] Question: Do the main claims made in the abstract and introduction accurately reflect the paper's contributions and scope?
    \item[] Answer: \answerYes{} 
    \item[] Justification: The claims in abstract and introduction are empirically supported in Sections 3 and 4 through experiments and filter visualizations.

\item {\bf Limitations}
    \item[] Question: Does the paper discuss the limitations of the work performed by the authors?
    \item[] Answer: \answerYes{} 
    \item[] Justification: The paper discusses limitations such as the need for further optimization of the 8 filters (Section 5).

\item {\bf Theory assumptions and proofs}
    \item[] Question: For each theoretical result, does the paper provide the full set of assumptions and a complete (and correct) proof?
    \item[] Answer: \answerNA{} 
    \item[] Justification: The paper is primarily empirical and does not include formal theorems or proofs, though it provides equations for filter approximation (e.g., Equation 1 and 2 in Section 3.1).

    \item {\bf Experimental result reproducibility}
    \item[] Question: Does the paper fully disclose all the information needed to reproduce the main experimental results of the paper to the extent that it affects the main claims and/or conclusions of the paper (regardless of whether the code and data are provided or not)?
    \item[] Answer: \answerYes{}{} 
    \item[] Justification: The paper details the models used, datasets, training durations (300 epochs), and evaluation metrics. It also outlines the autoencoder and greedy search method in detail (Sections 3.1 and 4.1).

\item {\bf Open access to data and code}
    \item[] Question: Does the paper provide open access to the data and code, with sufficient instructions to faithfully reproduce the main experimental results, as described in supplemental material?
    \item[] Answer: \answerNo{} 
    \item[] Justification: The paper uses public models and datasets. The scripts used in the paper will be available on a public Github repository after publication.

\item {\bf Experimental setting/details}
    \item[] Question: Does the paper specify all the training and test details (e.g., data splits, hyperparameters, how they were chosen, type of optimizer, etc.) necessary to understand the results?
    \item[] Answer: \answerYes{} 
    \item[] Justification:  The paper specifies training datasets, model sizes, training duration, and evaluation methods. In all experiments with each model, we used the exact same parameters sourced from original paper. See appendix.

\item {\bf Experiment statistical significance}
    \item[] Question: Does the paper report error bars suitably and correctly defined or other appropriate information about the statistical significance of the experiments?
    \item[] Answer: \answerNo{} 
    \item[] Justification: Since most experiments are performed on ImageNet, it would be computationally unaffordable to do multiple rounds of training. 

\item {\bf Experiments compute resources}
    \item[] Question: For each experiment, does the paper provide sufficient information on the computer resources (type of compute workers, memory, time of execution) needed to reproduce the experiments?
    \item[] Answer: \answerYes{} 
    \item[] Justification: See appendix.
    
\item {\bf Code of ethics}
    \item[] Question: Does the research conducted in the paper conform, in every respect, with the NeurIPS Code of Ethics \url{https://neurips.cc/public/EthicsGuidelines}?
    \item[] Answer: \answerYes{} 
    \item[] Justification: This work follows the NeurIPS Code of Ethics.

\item {\bf Broader impacts}
    \item[] Question: Does the paper discuss both potential positive societal impacts and negative societal impacts of the work performed?
    \item[] Answer: \answerNA{} 
    \item[] Justification: There is no societal impact of the work performed.
    
\item {\bf Safeguards}
    \item[] Question: Does the paper describe safeguards that have been put in place for responsible release of data or models that have a high risk for misuse (e.g., pretrained language models, image generators, or scraped datasets)?
    \item[] Answer: \answerNA{} 
    \item[] Justification: This work does not involve models or data with high risk of misuse, such as generative models or private data.

\item {\bf Licenses for existing assets}
    \item[] Question: Are the creators or original owners of assets (e.g., code, data, models), used in the paper, properly credited and are the license and terms of use explicitly mentioned and properly respected?
    \item[] Answer: \answerYes{} 
    \item[] Justification: All datasets and models used are cited.

\item {\bf New assets}
    \item[] Question: Are new assets introduced in the paper well documented and is the documentation provided alongside the assets?
    \item[] Answer: \answerYes{} 
    \item[] Justification: We present 8 master key filters and provide them in the appendix.

\item {\bf Crowdsourcing and research with human subjects}
    \item[] Question: For crowdsourcing experiments and research with human subjects, does the paper include the full text of instructions given to participants and screenshots, if applicable, as well as details about compensation (if any)? 
    \item[] Answer: \answerNA{} 
    \item[] Justification: The paper does not involve human subjects or crowdsourcing.

\item {\bf Institutional review board (IRB) approvals or equivalent for research with human subjects}
    \item[] Question: Does the paper describe potential risks incurred by study participants, whether such risks were disclosed to the subjects, and whether Institutional Review Board (IRB) approvals (or an equivalent approval/review based on the requirements of your country or institution) were obtained?
    \item[] Answer: \answerNA{} 
    \item[] Justification: No research involving human participants was conducted.

\item {\bf Declaration of LLM usage}
    \item[] Question: Does the paper describe the usage of LLMs if it is an important, original, or non-standard component of the core methods in this research? Note that if the LLM is used only for writing, editing, or formatting purposes and does not impact the core methodology, scientific rigorousness, or originality of the research, declaration is not required.
    \item[] Answer: \answerNA{} 
    \item[] Justification: This research does not involve LLMs as part of the core methodology.

\end{enumerate}

\end{document}